%% file: prelim.tex
\documentclass[runningheads]{llncs}
\usepackage{graphicx}

\usepackage[backend=bibtex, style=numeric, sortcites]{biblatex}
\bibliography{References.bib}

\usepackage{xcolor} %
\usepackage{amssymb} %
\usepackage{amsmath} %
\usepackage{enumitem} %
\usepackage[caption=false]{subfig}
\usepackage{subfloat}
\usepackage{algorithm}
\usepackage{algpseudocode}
\usepackage{booktabs} %
\usepackage[para]{threeparttable} %

\usepackage{xspace}
\newcommand{\bb}{\texttt{bb}\xspace}
\newcommand{\wb}{\texttt{wb}\xspace}
\newcommand{\gen}{\texttt{gen}\xspace}
\newcommand{\BB}{\texttt{BBA}\xspace}
\newcommand{\WB}{\texttt{WBA}\xspace}
\newcommand{\GEN}{\texttt{GENA}\xspace}
\newcommand{\PRELIM}{\texttt{PRELIM}\xspace}
\newcommand{\DT}{\texttt{DT}\xspace}
\newcommand{\CR}{\texttt{CR}\xspace}
\newcommand{\SD}{\texttt{SD}\xspace}
\newcommand{\NN}{\texttt{NN}\xspace}
\newcommand{\NNe}{\texttt{NNe}\xspace}
\newcommand{\SVM}{\texttt{SVM}\xspace}
\newcommand{\CRe}{\texttt{CRe}\xspace}
\newcommand{\RF}{\texttt{RF}\xspace}
\newcommand{\BT}{\texttt{BT}\xspace}
\newcommand{\RT}{\texttt{RT}\xspace}
\newcommand{\CART}{\texttt{CART}\xspace}
\newcommand{\RIPPER}{\texttt{RIPPER}\xspace}
\newcommand{\IREP}{\texttt{IREP}\xspace}
\newcommand{\BI}{\texttt{BI}\xspace}
\newcommand{\PRIM}{\texttt{PRIM}\xspace}
\newcommand{\DUMMY}{\texttt{DUMMY}\xspace}
\newcommand{\UNIF}{\texttt{UNIF}\xspace}
\newcommand{\NORM}{\texttt{NORM}\xspace}
\newcommand{\GMM}{\texttt{GMM}\xspace}
\newcommand{\GMMAL}{\texttt{GMMAL}\xspace}
\newcommand{\KDEM}{\texttt{KDEM}\xspace}
\newcommand{\KDE}{\texttt{KDE}\xspace}
\newcommand{\KDEB}{\texttt{KDEB}\xspace}
\newcommand{\CMM}{\texttt{CMM}\xspace}
\newcommand{\RERX}{\texttt{RE-RX}\xspace}
\newcommand{\VVA}{\texttt{VVA}\xspace}
\newcommand{\SMOTE}{\texttt{SMOTE}\xspace}
\newcommand{\ADASYN}{\texttt{ADASYN}\xspace}
\newcommand{\MUNGE}{\texttt{MUNGE}\xspace}
\newcommand{\SSL}{\texttt{SSL}\xspace}
\newcommand{\DTcv}{\texttt{DTcv}\xspace}
\newcommand{\DTcomp}{\texttt{DTcomp}\xspace}
\newcommand{\RR}{\texttt{RR}\xspace}
\newcommand{\XX}{\texttt{XX}\xspace}
\newcommand{\NO}{\texttt{NO}\xspace}
\newcommand{\DSIZE}{\lvert D\rvert}

\begin{document}
\title{Pedagogical Rule Extraction to Learn Interpretable Models~--- an Empirical Study
}
\author{Vadim Arzamasov \and
	Benjamin Jochum \and
	Klemens B\"ohm}
\authorrunning{V. Arzamasov et al.}
\institute{Karlsruhe Institute of Technology (KIT), Germany \\
	\email{\{vadim.arzamasov,klemens.boehm\}@kit.edu, uzebb@student.kit.edu}}
\maketitle              %
\begin{abstract}
Machine-learning models are ubiquitous.
In some domains, for instance, in medicine, the models' predictions must be interpretable. 
Decision trees, classification rules, and subgroup discovery are three broad categories of supervised machine-learning models presenting knowledge in the form of interpretable rules. 
The accuracy of these models learned from small datasets is usually low. 
Obtaining larger datasets is often hard to impossible.
Pedagogical rule extraction methods could help to learn better rules from small data by augmenting a dataset employing statistical models and using it to learn a rule-based model. 
However, existing evaluation of these methods is often inconclusive, and they were not compared so far.
Our framework \PRELIM unifies existing pedagogical rule extraction techniques.
In the extensive experiments, we identified promising \PRELIM configurations not studied before.

\keywords{Rule extraction \and Model explanation \and Density approximation \and Tabular data augmentation \and XAI.}
\end{abstract}

\input{sections/1_Introduction}
\input{sections/2_Related_Work}
\input{sections/3_Prelim_Idea}
\input{sections/4_Methods}

\input{sections/5_Experimental_Setup}

\input{sections/6_Results}
\input{sections/7_Discussion_and_FW}
\input{sections/8_Conclusion}

\printbibliography

\newpage
\appendix
\input{sections/Appendix}

\end{document}

%% file: sections/1_Introduction.tex
\section{Introduction}
Machine learning models are part of modern technology. 
While the internal structure of some models, e.g., artificial neural networks, is complex, other models, e.g., shallow decision trees, are relatively easy to understand for a human. 
We call them black-box (\bb) and white-box (\wb) respectively. 
\emph{Interpretability} is a desirable feature: 
The user of an interpretable model can validate its logic, decide whether to trust its predictions, and ensure the absence of discrimination. 
To achieve interpretability, one can (1) directly learn a white-box model or (2) train a black-box model and explain it~\cite{DBLP:journals/csur/GuidottiMRTGP19}. 
Some argue that the former option is preferable, at least for high-stakes decisions~\cite{rudin2019stop}. 
But learning an accurate white-box model tends to require large datasets that often are not available.

Pedagogical (black-box model agnostic) rule extraction (PRE) methods proposed in the last decades replace \bb model with a \wb model. %
They have sometimes been touted also as a means to improve white-box rule-based models learned from small datasets. %
Section~\ref{section:related_work} reviews these methods. %
However, it is unclear which methods improve the accuracy of which \wb models. %
This is because experiments in the respective publications do not tend to satisfy the following requirements. 

\textit{R1: White-Box Generality.}
Experiments should cover different categories of rule-based models. 
While decision tree (\DT) and classification rule (\CR) learners target at accurate classification, subgroup discovery (\SD) methods find individual rules covering many examples of a given class and possibly few examples of other classes~\cite{DBLP:conf/sigmod/ArzamasovB21}. 
Rule extraction literature has not covered all three categories at once. 

\textit{R2: Comparison to State-of-the-Art.}
To find the state-of-the-art, PRE methods should be compared. 
This has rarely been the case. Existing comparisons~\cite{domingos1997knowledge,DBLP:journals/tnn/FortunyM15} cover only few PRE methods implicitly. %

\textit{R3: Accuracy Increase.} We focus on increasing accuracy of a white-box model learned from small datasets, whereas many PRE methods exclusively aim at explaining a black-box model. 
In effect, PRE-related papers do not restrict the dataset size in the experiments and sometimes do not measure the accuracy~\cite{DBLP:journals/corr/BastaniKB17}. 

\textit{R4: Significance.}
A conclusive result requires experimenting with many datasets.
Many papers use fewer than seven datasets.

\textit{R5: White-Box Interpretability.}
Models obtained with \DT or \CR learners can loose interpretability if they consist of many rules; rules discovered with \SD methods are not interpretable if they have too many features in the antecedent~\cite{DBLP:conf/sigmod/ArzamasovB21}. Many related papers do not restrict the complexity of \wb models. 
This has sometimes resulted in models consisting of hundreds of rules~\cite{chen2004learning}. 

\textit{R6: Strong Baseline.}
Default hyperparameter values of a white-box model may cause overfitting. An overfitted model is a weak, easy-to-beat baseline. 
Although the importance of hyperparameter optimization has been demonstrated elsewhere~\cite{soenen2021effect}, many PRE papers do not claim to have this feature. 

\vspace{0.3em}
Starting point of this paper is \PRELIM (Pedagogical Rule Extraction for Learning Interpretable Models), a~framework unifying PRE methods to improve white-box rule-based models learned from small data. 
\PRELIM learns a ``generator'', a model that creates new data points, and a powerful black-box model that labels them. 
So \PRELIM creates extensive artificial data sets and uses them to train a \wb model.
With \PRELIM, we will demonstrate the practical importance of the above requirements. 
For instance, we have found that the majority of generators proposed for PRE so far, on average, do not increase the accuracy of a \wb model when its interpretability is controlled. %
We evaluate several generators never studied before with PRE methods %
and show that they help learning \wb models of higher quality.
It turns out that these generators are also competitive for explaining \bb models. In addition, we show the utility of \PRELIM for learning \wb models in a semi-supervised setting and for private data sharing.~--- Our code is openly available.\footnote{\url{https://github.com/Arzik1987/prelim}}

Paper outline: Section~\ref{section:related_work} is related work.
Section~\ref{section:methods} describes \PRELIM.
Section~\ref{section:experimental_setup} covers the experimental setup and says how it addresses the requirements. 
Section~\ref{section:results} features results. 
Section~\ref{section:future} describes future research. 
Section~\ref{section:conclusion} concludes.

%% file: sections/2_Related_Work.tex
\section{Related Work}
\label{section:related_work}
Related work is rule extraction, dataset augmentation, semi-supervised learning. %

\paragraph{Rule Extraction.}
Rule Extraction refers to the task of explaining the logic of a black-box model with a rule-based model~\cite{DBLP:journals/csur/GuidottiMRTGP19}. 
Rule extraction methods can be pedagogical or decompositional. 
The first group of methods treats a complex model as a black-box oracle~\cite{DBLP:journals/tnn/FortunyM15}, in the same way as \PRELIM does. %
The methods from this group differ from each other, mainly regarding the generators
and sometimes the white-box models they use.
Table~\ref{tab:related_work} reviews PRE methods and contrasts them with our work.

\begin{table}
	\centering
	\caption{Related methods. {\normalfont Column names: \BB: \bb model learner; \WB: \wb model learner; \textit{R2}: compared to related work; \textit{R3}: in the paper, the method has improved the \wb accuracy; \textit{R5}, \textit{R6}: evaluation complies with the respective requirement; \#D: the number of datasets used; \GEN: name of the most similar generator used in \PRELIM. Values: \RF: random forest; \NN: artificial neural network; \SVM: support vector machine; \BT: boosted trees, [{\XX}]\texttt{e}: an ensemble of models [{\XX}];  \CR/\RR: classification/regression rules; \DT/\RT: decision/regression tree; \SD: subgroup discovery; na: no information; $\infty$: explained in the text.}}
	\label{tab:related_work}
	\begin{threeparttable}
		\begin{tabular}{lccc|ccccccc}
			\toprule
			Name & Ref. & Year & \BB & \WB (\textit{R1}) &\textit{R2} & \textit{R3} & \#D (\textit{R4}) & \textit{R5} & \textit{R6} & \GEN \\ \hline
			noname & \cite{DBLP:conf/icml/CravenS94} & 1994 & \NN & \CR & {\color{lightgray}$\times$} & {\color{lightgray}na} & 1 & {\color{lightgray}$\times$} & {\color{lightgray}$\times$} & \UNIF \\
			
			Trepan & \cite{DBLP:conf/nips/CravenS95} & 1995 & \NN & \DT & {\color{lightgray}$\times$} & {\color{lightgray}na} & 4 & $\surd$ & {\color{lightgray}$\times$} & \KDEM %
			\\
			
			CMM & \cite{domingos1997knowledge} & 1997 & \CRe & \CR & {\color{lightgray}$\times$} & $\surd$ & 26 & {\color{lightgray}$\times$} & {\color{lightgray}$\times$} & \CMM \\
			
			noname & \cite{DBLP:journals/pr/KrishnanSB99} & 1999 & \NN & \DT & {\color{lightgray}$\times$} & {\color{lightgray}$\times$}\tnote{a} 
			& 2 & {\color{lightgray}$\times$} & {\color{lightgray}$\times$} & $\lnot$\VVA \\
			
			ANN-DT & \cite{DBLP:journals/tnn/SchmitzAG99} & 1999 & \NN & \RT & {\color{lightgray}$\times$} & {\color{lightgray}$\times$}\tnote{a} 
			& 3 & {\color{lightgray}$\times$}\tnote{b} & {\color{lightgray}$\times$} & \KDEB \\
			
			STARE & \cite{DBLP:conf/ijcnn/ZhouCC00} & 2000 & \NN & \CR & \cite{DBLP:conf/nips/CravenS95} & {\color{lightgray}na} & 6 & {\color{lightgray}$\times$} & {\color{lightgray}$\times$} & \UNIF \\
			
			noname & \cite{milare2001extracting} & 2001 & \NN & \DT, \CR & \cite{DBLP:conf/nips/CravenS95} & {\color{lightgray}$\times$} & 1 & {\color{lightgray}$\times$}\tnote{c}
			& {\color{lightgray}$\times$} & \DUMMY \\
			
			DecText & \cite{DBLP:conf/kdd/Boz02} & 2002 & \NN & \DT & {\color{lightgray}$\times$} & {\color{lightgray}na} & 3 & {\color{lightgray}$\times$} & {\color{lightgray}$\times$} & \KDEM \\
			
			REFNE & \cite{DBLP:journals/aicom/ZhouJC03} & 2003 & \NNe & \CR & {\color{lightgray}$\times$} & {\color{lightgray}na} & 6 & {\color{lightgray}$\times$} & {\color{lightgray}$\times$} & \UNIF \\
			
			REX & \cite{DBLP:conf/esann/Markowska-KaczmarT03} & 2003 & \NN & \CR & {\color{lightgray}$\times$} & {\color{lightgray}na} & 3 & $\surd$ & {\color{lightgray}$\times$} & \DUMMY \\
			
			BUR & \cite{chen2004learning} & 2004 & \SVM & \CR & {\color{lightgray}$\times$} & {\color{lightgray}na} & 2 & {\color{lightgray}$\times$}\tnote{c}
			& {\color{lightgray}$\times$} & \DUMMY \\
			
			Re-RX & \cite{DBLP:conf/icis/SetionoMB06} & 2006 & \NN & \CR
			& {\color{lightgray}$\times$} & {\color{lightgray}na} & 3 & {\color{lightgray}$\times$} & {\color{lightgray}$\times$} & \RERX \\
			
			ITER & \cite{DBLP:conf/dawak/HuysmansBV06} & 2006 & \SVM & \RR & {\color{lightgray}$\times$} & {\color{lightgray}$\times$} & 4
			& {\color{lightgray}$\times$} & {\color{lightgray}$\times$} & \DUMMY \\
			
			Minerva & \cite{DBLP:journals/tsmc/HuysmansSBV08} & 2008 & \SVM & \CR & \cite{DBLP:conf/nips/CravenS95} & {\color{lightgray}na} & 8
			& {\color{lightgray}$\times$} & {\color{lightgray}$\times$}
			& \DUMMY \\
			
			GPDT & \cite{DBLP:conf/cidm/JohanssonN09} & 2009 & \NNe & \DT & {\color{lightgray}$\times$} & $\surd$ & 26 & {\color{lightgray}na} & {\color{lightgray}$\times$} & \DUMMY \\
			
			RF2TREE & \cite{DBLP:journals/spe/JiangLZ09} & 2009 & \RF & \DT & {\color{lightgray}$\times$} & $\surd$ & 24 & {\color{lightgray}na} & {\color{lightgray}$\times$} & \NORM \\
			
			CAD-MDD & \cite{gibbons2013computerized} & 2014 & \RF & \DT & {\color{lightgray}$\times$} & {\color{lightgray}na} & 1 & $\surd$ & {\color{lightgray}$\times$} & {\color{lightgray}na} \\
			
			ALPA & \cite{DBLP:journals/tnn/FortunyM15} & 2015 & \RF, \SVM, \NN & \DT, \CR & {\color{lightgray}$\times$} & $\surd$ & 25 & {\color{lightgray}$\times$}\tnote{b}
			& {\color{lightgray}$\times$} & \VVA \\
			
			STA & \cite{zhou2016interpreting} & 2016 & \RF & \DT & {\color{lightgray}$\times$} & {\color{lightgray}na} & 2 & $\surd$ & {\color{lightgray}$\times$} & \KDEM \\
			
			noname & \cite{DBLP:journals/corr/BastaniKB17} & 2017 & \RF, \NN & \DT & {\color{lightgray}$\times$} & {\color{lightgray}na} & 8 & $\surd$ & {\color{lightgray}$\times$} & \GMMAL \\
			
			REDS & \cite{DBLP:conf/sigmod/ArzamasovB21} & 2021 & \RF, \BT, \SVM & \SD & {\color{lightgray}$\times$} & $\surd$ & 33 & $\surd$ & $\surd$ & \UNIF \\
			
			\PRELIM & our & 2021 & \RF, \BT & \DT, \CR, \SD & $\infty$ & $\surd$ & 30 & $\surd$ & $\surd$ & $\infty$ \\ \hline
		\end{tabular}
		\begin{tablenotes}\footnotesize
			\item[a]{Accuracy increases on one dataset only.}
			
			\item[b]{Complexity is set similar to the one of the baseline that was not controlled.}
			
			\item[c]{The complexity restriction is too soft and allows \wb models with $>100$ rules.}
		\end{tablenotes}
	\end{threeparttable}
\end{table}

Related work does not satisfy the requirements listed in the introduction. 
In particular, only~\cite{DBLP:conf/ijcnn/ZhouCC00,milare2001extracting,DBLP:journals/tsmc/HuysmansSBV08} explicitly compare their methods to another PRE algorithm. 
Thus, related work generally does not satisfy \textit{R2}. 
We use various generators proposed for pedagogical rule extraction and other generators with \PRELIM, hence the symbol $\infty$ in the column \textit{R2} in Table~\ref{tab:related_work}. 
The column ``\GEN'' identifies the generator we test with \PRELIM that is most similar to the one proposed for the corresponding PRE method. 

Only~\cite{domingos1997knowledge,DBLP:conf/cidm/JohanssonN09,DBLP:journals/spe/JiangLZ09,DBLP:journals/tnn/FortunyM15,DBLP:conf/sigmod/ArzamasovB21} address \textit{R3}. 
However, \cite{domingos1997knowledge,DBLP:conf/cidm/JohanssonN09,DBLP:journals/spe/JiangLZ09,DBLP:journals/tnn/FortunyM15}, do not restrict the complexity of a white-box model, so they do not comply with \textit{R5}.

All papers, except for \cite{DBLP:conf/sigmod/ArzamasovB21}, do not satisfy \textit{R6}. 
Although~\cite{DBLP:conf/sigmod/ArzamasovB21} complies with both \textit{R6} and \textit{R5}, it covers only datasets produced in computer experiments where one knows the distribution of inputs and does not need to develop a generator approximating it.

None of the papers considers three categories of rule-based \wb model learners \DT, \CR, and \SD. 
So they do not address \textit{R1}. Much related work uses few datasets for experiments~--- it does not satisfy \textit{R4}.

\paragraph{Dataset Augmentation.}
Rule extraction techniques are part of a larger group of knowledge distillation methods~\cite{DBLP:journals/corr/HintonVD15} where the end model is generally not interpretable. 
One of the generators we use in \PRELIM comes from~\cite{DBLP:conf/kdd/BucilaCN06} that falls into this group. 

Knowledge distillation techniques in turn often rely on dataset augmentation methods that create artificial data and add it to the train set. 
They are proven particularly effective for object or speech recognition tasks where invariant (producing objects of the same class) input transformations are apparent. 
For instance, such transformations are scaling or shifting images~\cite{Goodfellow-et-al-2016}. 

Several data augmentation techniques exist for tabular data, e.g.,~\cite{DBLP:conf/nips/FakoorMECS20,DBLP:journals/corr/abs-1811-11264}. To our knowledge, they either have not been used for training white-box models~\cite{DBLP:conf/nips/FakoorMECS20} or did not increase their quality~\cite{DBLP:journals/corr/abs-1811-11264}.

\paragraph{Semi-Supervised Learning.}
In the presence of unlabeled data, some semi-supervised techniques use a metamodel to create pseudo-labels and augment the train set~\cite{DBLP:conf/cvpr/XieLHL20,zhu2005semi}. Our setting is more general as we do not assume the existence of unlabeled data. 

%% file: sections/3_Prelim_Idea.tex
\section{PRELIM}
\label{section:methods}

This section describes the specifics of \PRELIM, the evaluation framework used here. %
We start with notation.
Let $D^\textit{tr}$ be a train dataset:
$$D^\textit{tr}=\begin{pmatrix}
	x_{11} &  \dots & x_{1M} & y_1 \\
	\vdots &  \ddots & \vdots & \vdots \\
	x_{N1} & \dots & x_{NM} & y_N \\
\end{pmatrix}.$$
In each row $i\in\{1,\dots,N\}$, the first $M$ elements contain feature values $x_i=\left(x_{i1}\dots,x_{iM}\right)$ and constitute a \emph{point} in an $M$-dimensional space; the last column $\left(y_1,\dots,y_N\right)$ contains the target class for each point. This work considers a binary classification problem with continuous features, i.e., $x_i\in\mathbb{R}^M$, $y_i\in\{0,1\}$. 
We refer to the entire row $d_i=\left(x_i,y_i\right)$ as an \emph{example}. 

\begin{algorithm}[t]
	\caption{\PRELIM}\label{alg:PRELIM}
	\begin{algorithmic}[1]
		\Procedure{\PRELIM}{$D^\textit{tr},{\BB,\WB,\GEN},L,\dots$}
		\State ${\bb} = \BB.\texttt{fit}(D^\textit{tr},\dots)$
		\State $\gen = \texttt{\GEN.fit}(D^\textit{tr},\dots)$
		\State $D^\textit{new}=[]$
		\For{$0 < i < L+1$}
		\State $x_i^\textit{new}=\texttt{gen.sample}()$
		\State $y_i^\textit{new}=\texttt{bb.predict}(x_i^\textit{new})$
		\State $D^\textit{new}.\texttt{append}(x_i^\textit{new},y_i^\textit{new})$
		\EndFor
		\State $\texttt{wb} = \texttt{WBA.fit}(D^\textit{tr}\cup D^\textit{new},\dots)$
		\State \textbf{return} $\texttt{wb}$
		\EndProcedure
	\end{algorithmic}
\end{algorithm}

\subsection{The Algorithm}
Algorithm~\ref{alg:PRELIM} is the \PRELIM framework. %
It takes as input a dataset $D^\textit{tr}$, a black-box model learning algorithm $\BB$, a white-box model learning algorithm $\WB$, %
a generator learning algorithm $\GEN$, and the number $L$ of examples to create. It outputs a white box model $\wb$. \PRELIM works as follows.
\begin{enumerate}[noitemsep]
	\item Use $D^\textit{tr}$ with algorithms $\BB$ and $\GEN$ to obtain a black-box model $\bb$ and generator $\gen$ (Lines~2--3); 
	\item create $L$ artificial points with $\gen$ and label them using $\bb$ to form a new dataset $D^\textit{new}$.
	\item use $D^\textit{tr}\cup D^\textit{new}$ with algorithm $\WB$ to get a white-box model $\wb$
\end{enumerate}

\PRELIM differs from the algorithms in Table~\ref{tab:related_work} in that it does not rely on a particular generator but rather takes one as input. This renders \PRELIM more flexible and allows to experiment with different generators.

Next, we describe the white-box models, the black-box models, and the generators we employ. 
Our experiments compare various generators and suggest promising combinations of $\GEN$ and $\BB$. %
\PRELIM is easy to extend and lets its users add new components.

%% file: sections/4_Methods.tex
\subsection{White-Box Models}
With \PRELIM, we use five white-box models of three categories: decision trees, classification rules, and subgroup discovery. 
Decision trees and classification rules aim at maximizing accuracy. 
Subgroup discovery methods search for large groups of examples of the same class, they use weighted relative accuracy (WRAcc) quality measures. 
Section~\ref{app:wbmodels} explains the three groups in detail\footnote{Sections named with letters are in the supplementary material.}; \cite{DBLP:series/cogtech/FurnkranzGL12} describes similarities and differences between these models.

We use a version of the decision tree learning algorithm, \CART~\cite{DBLP:books/wa/BreimanFOS84},
a rule learning algorithm {\IREP}~\cite{DBLP:conf/icml/FurnkranzW94}, and its successor {\RIPPER}~\cite{DBLP:conf/icml/Cohen95}, subgroup discovery algorithms \PRIM~\cite{DBLP:journals/sac/FriedmanF99}, and \BI~\cite{DBLP:conf/icdm/MampaeyNFK12}. The implementations of \CART, \RIPPER, \IREP, and \BI allow restricting the complexity of the learned model, as we will explain. %

\subsection{Generators}

We describe two groups of generators. The first includes sampling algorithms that have been used for PRE and extensions of these algorithms we propose. %
The second group contains generators that have not been used with PRE methods. %

\paragraph{Conventional Generators and Extensions.}
Literature (Table~\ref{tab:related_work}) uses several algorithms for PRE. We describe them below and propose extensions. %

\paragraph{\DUMMY.}  Similarly to~\cite{milare2001extracting,DBLP:conf/esann/Markowska-KaczmarT03,chen2004learning,DBLP:conf/dawak/HuysmansBV06,DBLP:journals/tsmc/HuysmansSBV08,DBLP:conf/cidm/JohanssonN09}, \DUMMY returns all points from $D$.

\paragraph{\UNIF.} This generator creates new points by sampling the value of each feature, i.i.d.\ from a continuous uniform distribution with bounds defined by minimum and maximum values of the feature in $D$. References~\cite{DBLP:conf/icml/CravenS94,DBLP:conf/ijcnn/ZhouCC00,DBLP:journals/aicom/ZhouJC03,DBLP:conf/sigmod/ArzamasovB21} do similarly.

\paragraph{\NORM.} As~\cite{DBLP:journals/spe/JiangLZ09} does, \NORM generates points by sampling the value of each feature, i.i.d.\ from a Gaussian distribution with mean and standard deviation estimated from the values of the feature in $D^\textit{tr}$.

\paragraph{\GMM.} Gaussian mixture models represent a probability density function (pdf) as a weighted mixture of $k$ Gaussian distributions~\cite{DBLP:books/lib/HastieTF09}: %
	$$f(x)=\sum_{i=1}^{k}\alpha_i\mathcal{N}\left(\mu_i,\Sigma_i\right).$$
\GMM uses the resulting composite pdf to generate new points.
The weights $\alpha_i$ and the elements of the covariance matrix $\Sigma_i$ are estimated from the data. 
The structure of the covariance matrix and the number of components $k$ are hyperparameters of \GMM. %

\paragraph{\GMMAL} This generator from~\cite{DBLP:journals/corr/BastaniKB17} refers to the Gaussian mixture model with the diagonal covariance matrix.

\paragraph{\KDEM.} 
Kernel density estimation~\cite{DBLP:books/sp/Silverman86} is another method to approximate pdfs. 
For this method, one has to specify a kernel function and a bandwidth parameter. 
For \KDEM, in line with~\cite{DBLP:conf/nips/CravenS95,DBLP:conf/kdd/Boz02,zhou2016interpreting}, we use a Gaussian kernel and model the distribution of each feature separately. 

\paragraph{\KDE.} This generator extends \KDEM to a multidimensional case. 
For \KDE, one specifies a set of bandwidths rather than a single value.

\paragraph{\KDEB} The ANN-DT algorithm~\cite{DBLP:journals/tnn/SchmitzAG99} suggests to sample points uniformly at random from $M$-balls with a radius $r$ and centered at points of the train data. One can see \KDEB as a variant of \KDE, akin to nearest-neighbor density estimation~\cite{DBLP:books/wi/Scott92}.

\paragraph{\CMM.}
The CMM method~\cite{domingos1997knowledge} uses an opaque model, C4.5-rules ensemble, to generate artificial points and to label them. 
It creates new points by sampling from the area defined by each rule in the ensemble uniformly at random. 
The number of artificial points is proportional to the number of points in train data covered with the respective rule. We use a similar approach, \CMM, with a random forest instead of a C4.5-rules ensemble.

\paragraph{\RERX.} The \RERX method~\cite{DBLP:conf/icis/SetionoMB06} uses a subset of train examples where $\bb$ predictions match true labels.

\paragraph{\VVA.} The ALPA algorithm~\cite{DBLP:journals/tnn/FortunyM15} creates new points that are (a) not very different from points in train data and (b) lie near the decision boundary of a $\bb$ to be explained. Section~\ref{app:vva} describes this generator. 
\cite{DBLP:journals/pr/KrishnanSB99} proposes a generator doing almost the opposite; we do not use it due to an unclear description.

\paragraph{Other generators.}
To deal with the imbalanced classification problem, one sometimes uses over-sampling techniques that create artificial examples of the minority class. 
We turn these methods into generators for \PRELIM by constructing an artificial classification problem. To do so, we assume that points from $D^\textit{tr}$ belong to the minority class and generate points from the majority class with \UNIF generator described before. %
We adapted the \SMOTE and \ADASYN generators in this way.

\paragraph{\SMOTE.}
SMOTE~\cite{DBLP:journals/jair/ChawlaBHK02} (Synthetic Minority Over-sampling Technique) takes the train data belonging to the minority class and creates new data for each point $x_i$ as follows.
First, it finds the $k$ nearest neighbors of $x_i$  belonging to the minority class and selects one of them, $x_i'$. 
It then creates a new point $x_i^\textit{new}$ with coordinates $x_{ij}^\textit{new}=\textit{gap}_{ij}\cdot x_{ij} + (1-\textit{gap}_{ij})x'_{ij}$, where $\textit{gap}_{ij}$ is a random number between 0 and 1, new for each point, coordinate and iteration. 
SMOTE stops after creating a new dataset of the required size. 
The number of nearest neighbors $k$ is a hyperparameter of SMOTE; \SMOTE is the corresponding \PRELIM generator.

\paragraph{\ADASYN.} ADASYN~\cite{DBLP:conf/ijcnn/HeBGL08} creates more synthetic points than SMOTE for the minority class points densely surrounded by the majority class examples. 
The respective \PRELIM generator is \ADASYN.

\paragraph{\MUNGE.} The authors of~\cite{DBLP:conf/kdd/BucilaCN06} replace an ensemble of artificial neural networks with a shallow neural network. We use their algorithm \MUNGE to generate new points. 
For each point $x_i$ in train, it finds its nearest neighbor $x_i'$. Then \MUNGE creates a new point by changing each coordinate $x_{ij}$ of $x_i$, $j=1,\dots,M$ with a predefined probability $P$ to a random value sampled from the Gaussian distribution $\mathcal{N}\left(x_{ij},\lvert x_i-x_i'\rvert /s\right)$, $P$, $s$ are hyperparameters of \MUNGE. 
One can see \MUNGE as a random-bandwidth variant of \KDE, similar to nearest neighbor density estimation~\cite{DBLP:books/wi/Scott92}. \MUNGE is also similar to \SMOTE with $k=1$.

\paragraph{\SSL.} To test \PRELIM in a semi-supervised learning setting, we introduce the \SSL generator; it returns some existing points from test data without their labels.

\subsection{Black-Box Models.}
We use random forests (\RF)~\cite{DBLP:journals/ml/Breiman01} and boosted trees (\BT)~\cite{DBLP:conf/kdd/ChenG16} as black-box learners (\BB) since they perform well in classification tasks~\cite{DBLP:journals/jmlr/WainbergAF16} and for rule extraction~\cite{DBLP:conf/sigmod/ArzamasovB21}. Beyond label predictions, both models can output continuous probability scores that are essential for the \VVA generator; we also use the scores in the experiments with subgroup discovery.

%% file: sections/5_Experimental_Setup.tex
\section{Experimental Setup}
\label{section:experimental_setup}
This section presents our experimental setup and explains how it satisfies the requirements \textit{R1}--\textit{R6}.
Section~\ref{app:experimental_setup} provides further details on implementation, datasets, and hyperparameter values for $\BB$ algorithms and generators.

\subsection{Datasets}
We use 30 datasets from the UCI~\cite{Dua:2019}, PMLB~\cite{romano2021pmlb} and OpenML~\cite{DBLP:journals/sigkdd/VanschorenRBT13} repositories. 
We removed categorical features, features taking fewer than 20 unique values, and rows with missing values. We denote each resulting dataset with $D$. Using many datasets allows us to satisfy \textit{R4: Significance}. %

\subsection{Design of Experiments}
\label{section:design_of_experiments}
A single experiment has the following steps
\begin{enumerate}[noitemsep]
	\item Split a dataset into $D^\textit{tr}$ with $\lvert D^\textit{tr}\rvert=N$ and $D^\textit{test}$; %
	\label{item:split}
	\item normalize features in $D^\textit{tr}$, transform $D^\textit{test}$ correspondingly; \label{item:scale}
	\item obtain $\wb=\PRELIM(D^\textit{tr},\BB,\WB,\GEN,L,\dots)$; %
	\label{item:PRELIM}
	\item evaluate quality of the resulting model $\wb$ on $D^\textit{test}$. %
	\label{item:quality}
\end{enumerate}
We experiment with the generators and black-box and white-box model learners listed in Section~\ref{section:methods}. For each dataset, we experiment with $N=\lvert D^\textit{tr}\rvert \in\{100,\allowbreak 400\}$.
To average random effects, we do $K=25$ splits into $D^\textit{tr}$ and $D^\textit{test}$; sets $D^\textit{tr}$ resulting from different splits overlap as little as possible. %

\subsection{Quality Measures}
The motivation behind \PRELIM is to increase the accuracy of decision trees and classification rules or the WRAcc of subgroups. 
$$
\textrm{WRAcc}=\frac{n}{N}\left(\frac{n^+}{n}-\frac{N^{+}}{N}\right),
\label{eq:wracc}
$$
where $n$, $n^+$, are the total number of examples satisfying conditions in the rule antecedent and the sum of their $y$ values, respectively; $N$, $N^+$ are the corresponding values for the whole dataset. %
Since some of datasets are imbalanced, one is also interested in balanced accuracy (BA), an average value of true positive rate and true negative rate.
Measuring accuracy, BA, WRAcc, and comparing them to the baseline (Step~\ref{item:PRELIM} in Section~\ref{section:design_of_experiments}) satisfies \textit{R3: Accuracy Increase}.
For decision trees and classification rules, we also evaluate how well \PRELIM explains a black-box model. 
The respective measure is fidelity, the accuracy of a white-box model over black-box model predictions rather than over true labels~\cite{DBLP:journals/tnn/SchmitzAG99}. %
We count the number of leaves in decision trees, the number of rules in decision lists of \RIPPER and \IREP, and the number of features in the rule antecedent in subgroups. 
These are conventional proxies of white-box interpretability~\cite{DBLP:journals/tnn/FortunyM15}. 

\subsection{Hyperparameters}
\label{section:hyperparameters}
We use grid search hyperparameter optimization with 5-fold cross-validation unless otherwise specified.  %
Using various white-box models and generators satisfies \textit{R1: White-Box Generality} and \textit{R2: Comparison to State-of-the-Art}.

Varying hyperparameters of white boxes lets our experiments comply with \textit{R5: White-Box Interpretability} and \textit{R6: Strong Baseline}. 
We optimize hyperparameters of white-box learners using $D^\textit{tr}$. 

\paragraph{White-Box Models.}
We experiment with three parametrizations of the decision tree learning algorithm, dubbed \DTcomp, \DTcv, and \DT. %
For \DT, we restrict the number of samples reaching non-leaf nodes to be greater than $10$, to limit its depth somewhat. %
Still, \DT does not satisfy neither \textit{R5} nor \textit{R6}
and we use it to show the importance of these requirements. 
For \DTcomp, we limit the number of leaves in the tree to eight; %
\DTcomp thus complies with \textit{R5}. \DTcv obeys \textit{R6}, we optimize the number of leaves,
choosing it from $2^{\{1,2,3,4,5,6,7\}}$. %
For a \DTcv model obtained with \PRELIM, we ensure that it does not have more leaves than \DTcv learned from $D^\textit{tr}$, so that the accuracy comparison is fair. %

For \IREP and \RIPPER, we limit the number of rules they output to eight, to comply with \textit{R5}. 

In preliminary experiments, we have found \PRELIM with subgroup discovery methods to work better if a black-box model assigns class probabilities rather than class labels to examples in $D^\textit{tr}\cup D^\textit{new}$; we set up \PRELIM accordingly. %
Next, we adjust \BI and \PRIM to find precisely one subgroup; this simplifies the analysis and does not reduce generality of the result~\cite{DBLP:conf/sigmod/ArzamasovB21}. %

We limit the number of features in the rule antecedent for \BI to 15 and select it from five variants $\{Z-j\lceil Z/5\rceil\}$, $j>0$, $j\lceil Z/5\rceil<Z$, $Z=\min(15,M)$. %
We ensure that the \BI algorithm with \PRELIM is not exposed to more features than \BI applied solely to~$D^\textit{tr}$. \BI satisfies \textit{R5} and \textit{R6}. %
We have also optimized a hyperparameter of \PRIM (cf.\ Section~\ref{app:experimental_setup}), so \PRIM complies with \textit{R6}.

\paragraph{\PRELIM.}
Some generators automatically determine the size $L$ of the generated sample. 
In particular, \DUMMY creates a sample of size $L=N$, in \RERX $L\le N$, \VVA internally optimizes the ratio $L/N$. 
For \SSL, we use $L=\min(10^4-N,\lfloor(\DSIZE-N)/2\rfloor)$ points without their labels from $D^\textit{test}$, so that at least half of $D^\textit{test}$ remains for testing. %
For other generators, $L=10^5-N$ if the $\WB$ is a decision tree learner and $L=10^4-N$ otherwise. Large $L$ values tend to be better~\cite{DBLP:conf/sigmod/ArzamasovB21}. We limit $L$ to keep the runtime reasonable.

When balanced accuracy is of interest, we adjust hyperparameters of \BB and \WB in \PRELIM to assign higher weights to the minority class.

%% file: sections/6_Results.tex
\section{Results}
\label{section:results}

We present the results separately for decision trees, classification rules, and subgroup discovery. 
There is a heat map for each white-box model, dataset size $N$, and quality measure. %
Each cell within a heat map is the quality estimate averaged across $30\times K$ experiments for each black-box model (columns) and generator (rows) combination. 
Grouped columns refer to different \wb models. %
Here, 30 is the number of datasets, and $K$ is the number of repetitions, cf.\ Section~\ref{section:design_of_experiments}.  
The generator \NO stands for the baseline experiments, i.e., training the \wb model solely from~$D^\textit{tr}$. 
Blue cells stand for a quality increase over the baseline; higher saturation means greater improvement. 
Yellow cells stand for quality drops. 
We present results for the semi-supervised learning setting (\SSL generator) together with the others but discuss them separately. %

An average value can summarize few experiments with substantially different quality. 
Hence, we additionally report the number of wins, draws, and losses of \PRELIM over the baseline for selected \PRELIM instantiations and quality measures.

\paragraph{Decision Tree.} %

The accuracy of the na\"ive classifier that always predicts the majority class in $D^\textit{tr}$ is between 50\% (for perfectly balanced data) and 100\%. 
In datasets we use the classes often are not perfectly balanced. 
To account for this, %
we report the relative accuracy increase, i.e., the difference between the accuracy of the model \wb and of the na\"ive classifier. 
Similarly, we report a relative fidelity increase and {relative balanced accuracy increase}. 

\begin{figure}[t]
	\centering
	{\includegraphics[width=0.51\columnwidth]{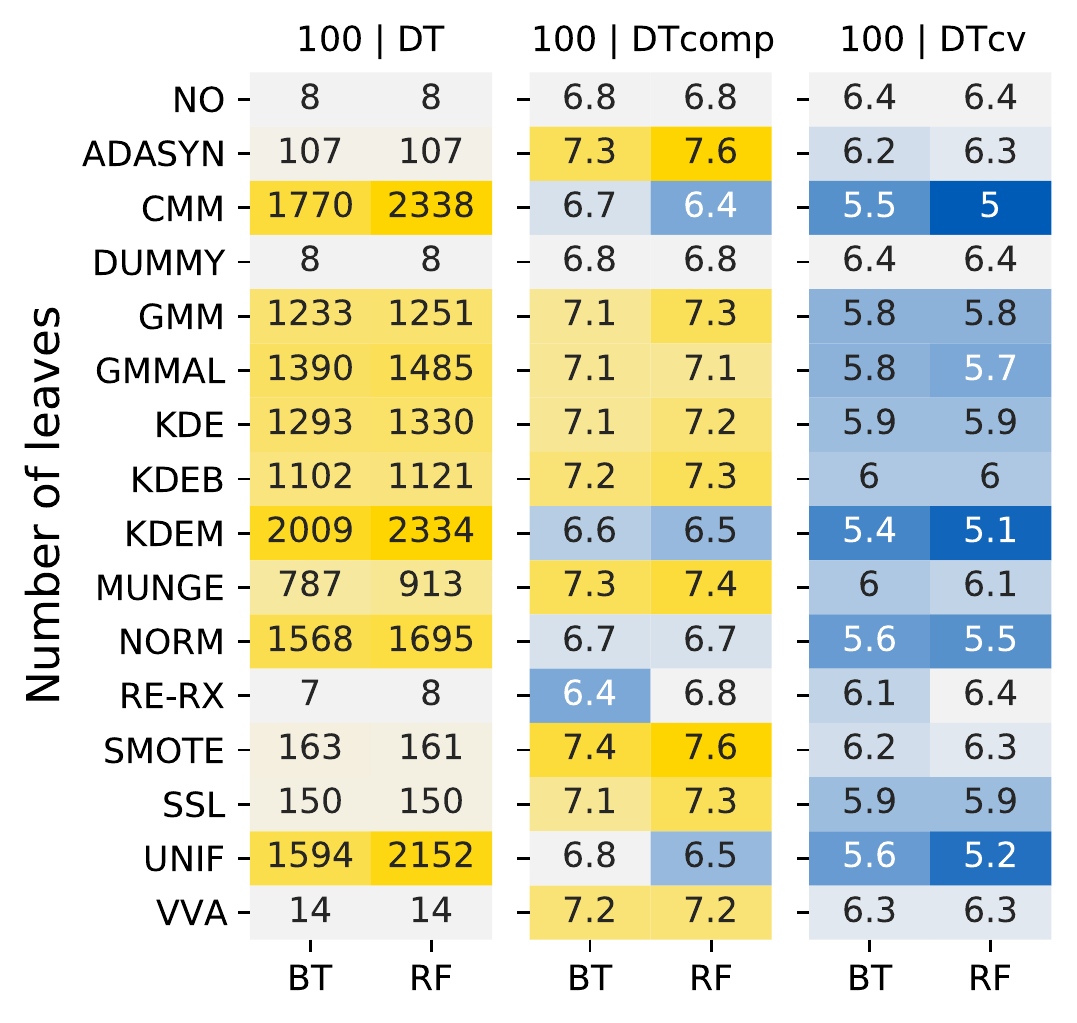}}
	\quad
	{\includegraphics[width=0.40\columnwidth]{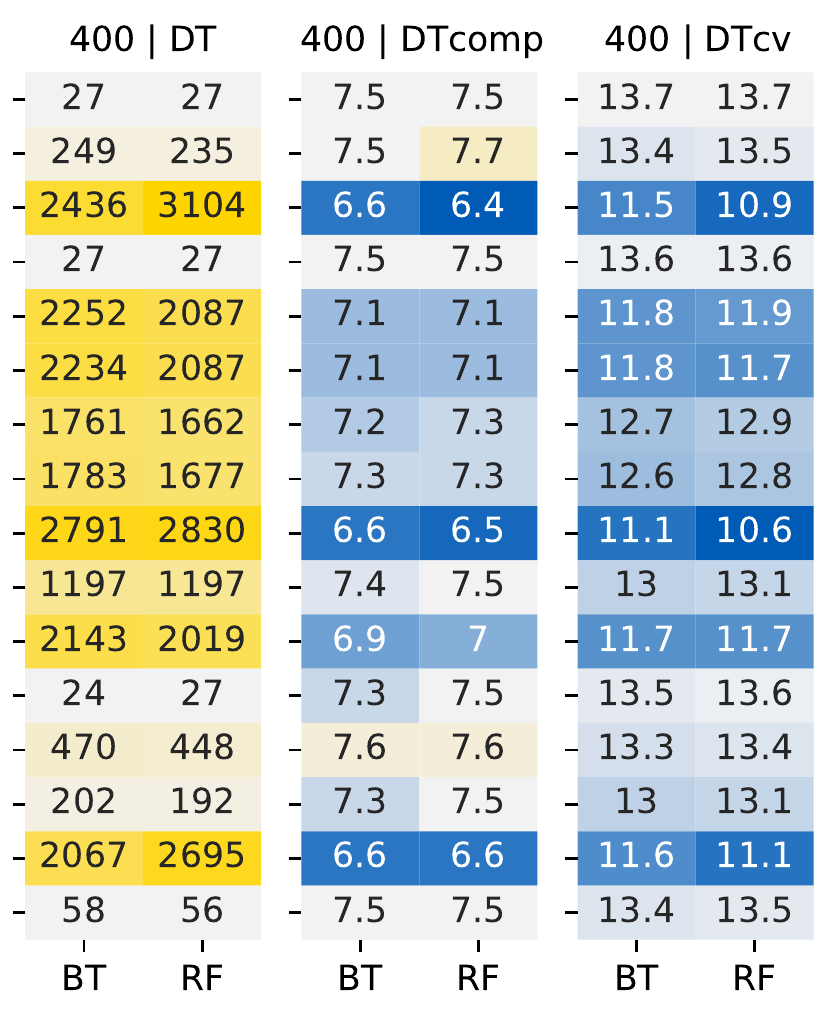}} 
	\caption{Decision tree. Average number of leaves for $N=100$ (left) and $N=400$ (right)}
	\label{fig:dt_nle}
\end{figure}

\begin{figure}
	\centering
	{\includegraphics[width=0.51\columnwidth]{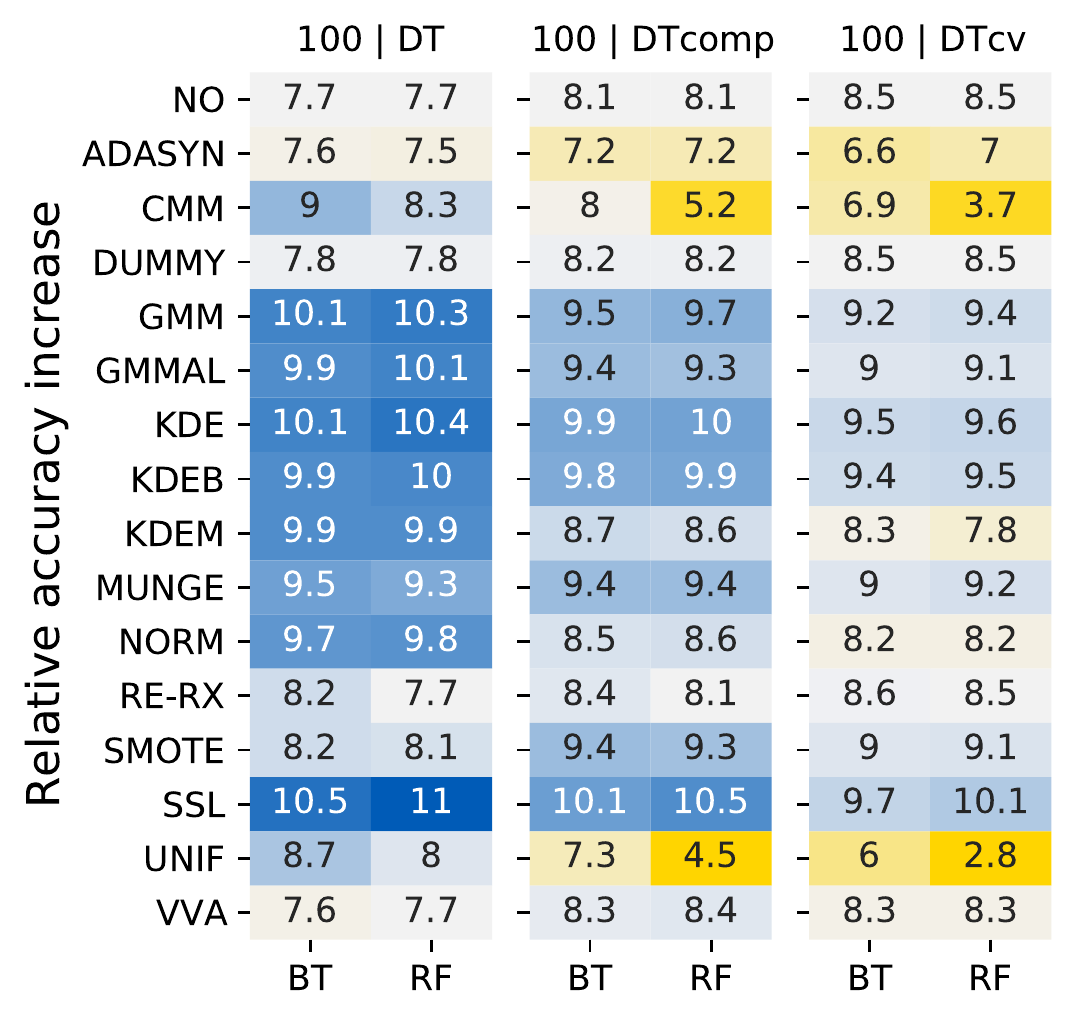}}
	\quad
	{\includegraphics[width=0.40\columnwidth]{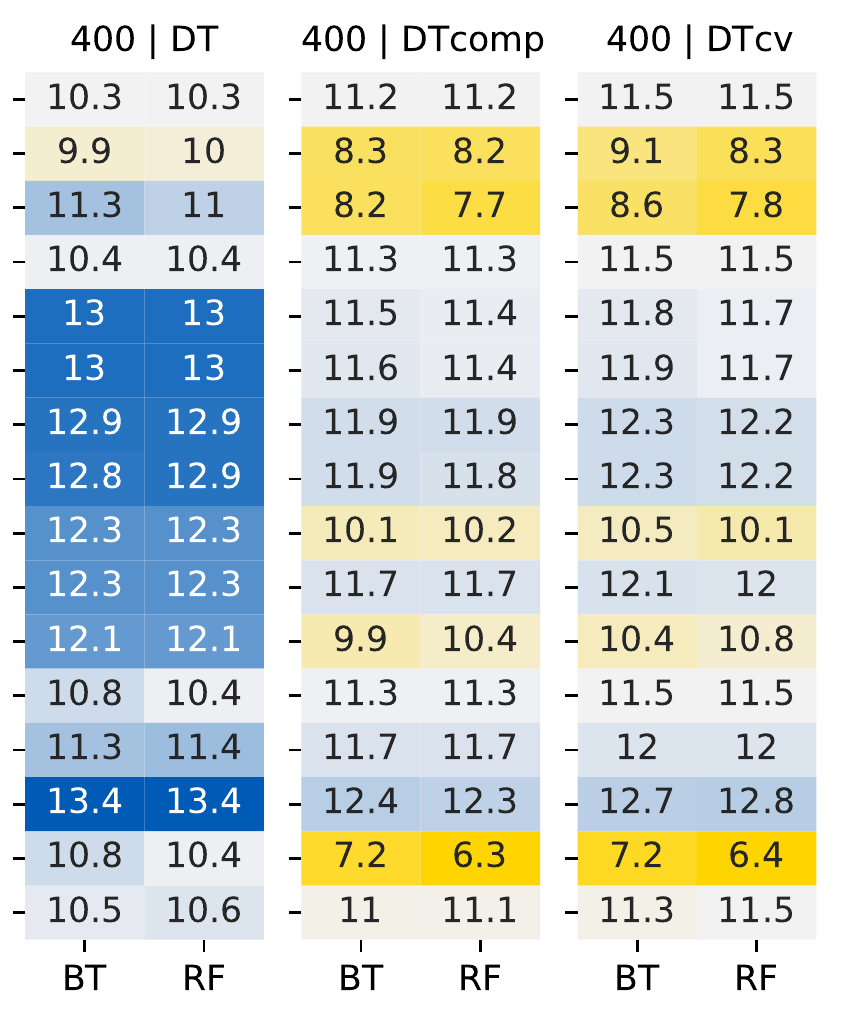}} 
	\\
	{\includegraphics[width=0.51\columnwidth]{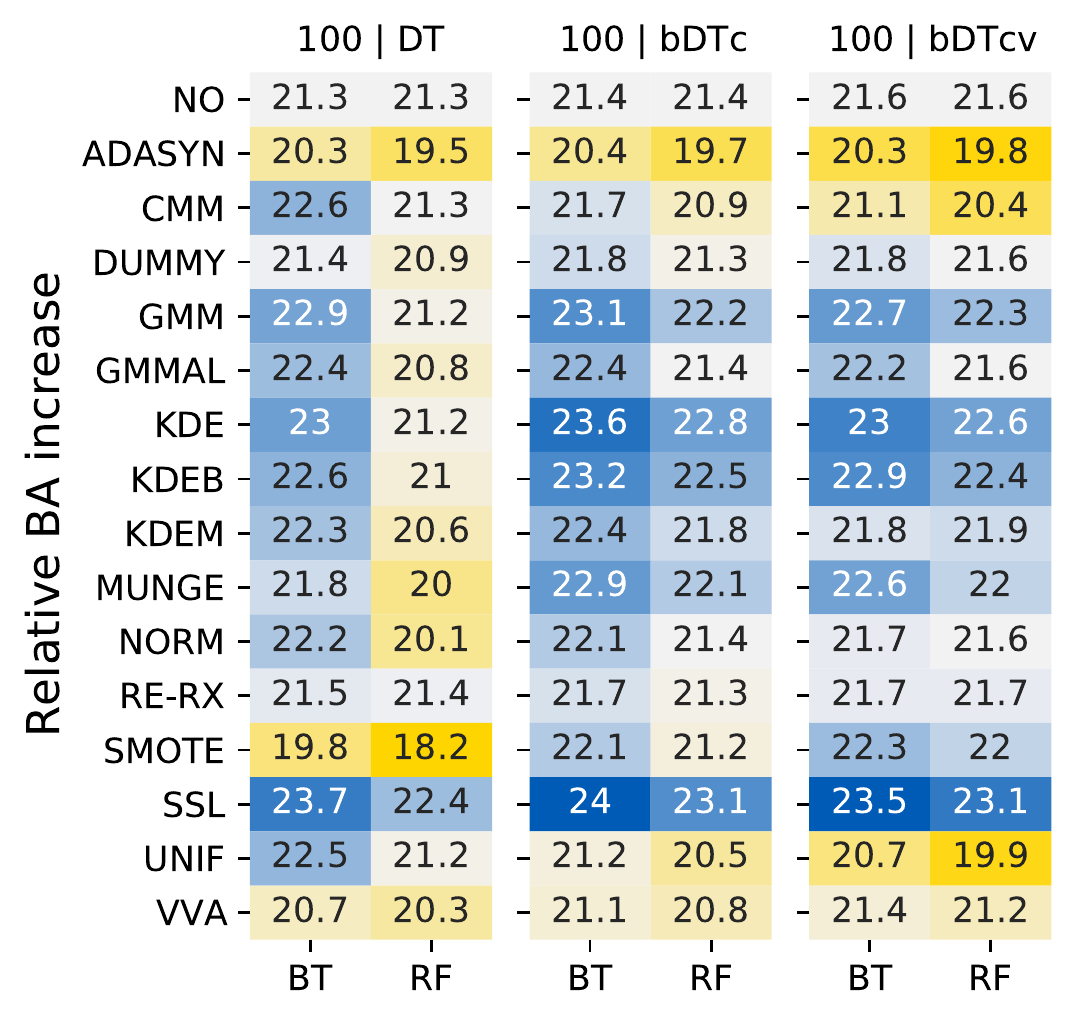}}
	\quad
	{\includegraphics[width=0.40\columnwidth]{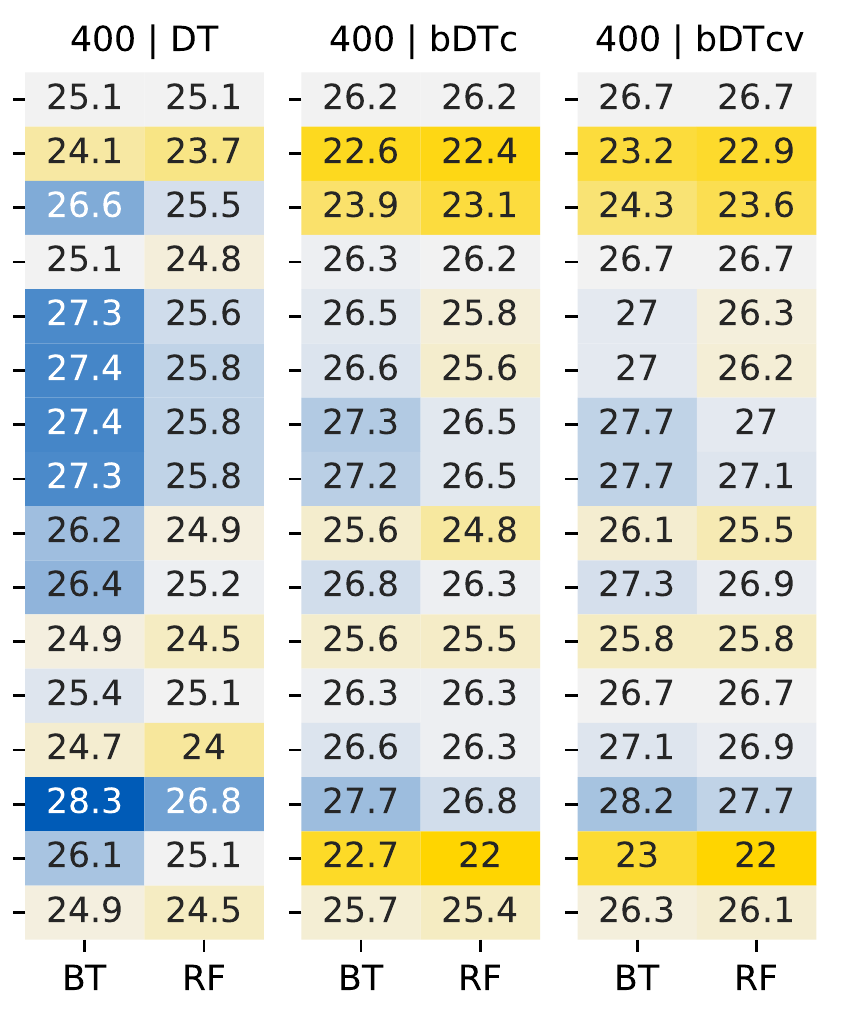}} 
	\\
	{\includegraphics[width=0.51\columnwidth]{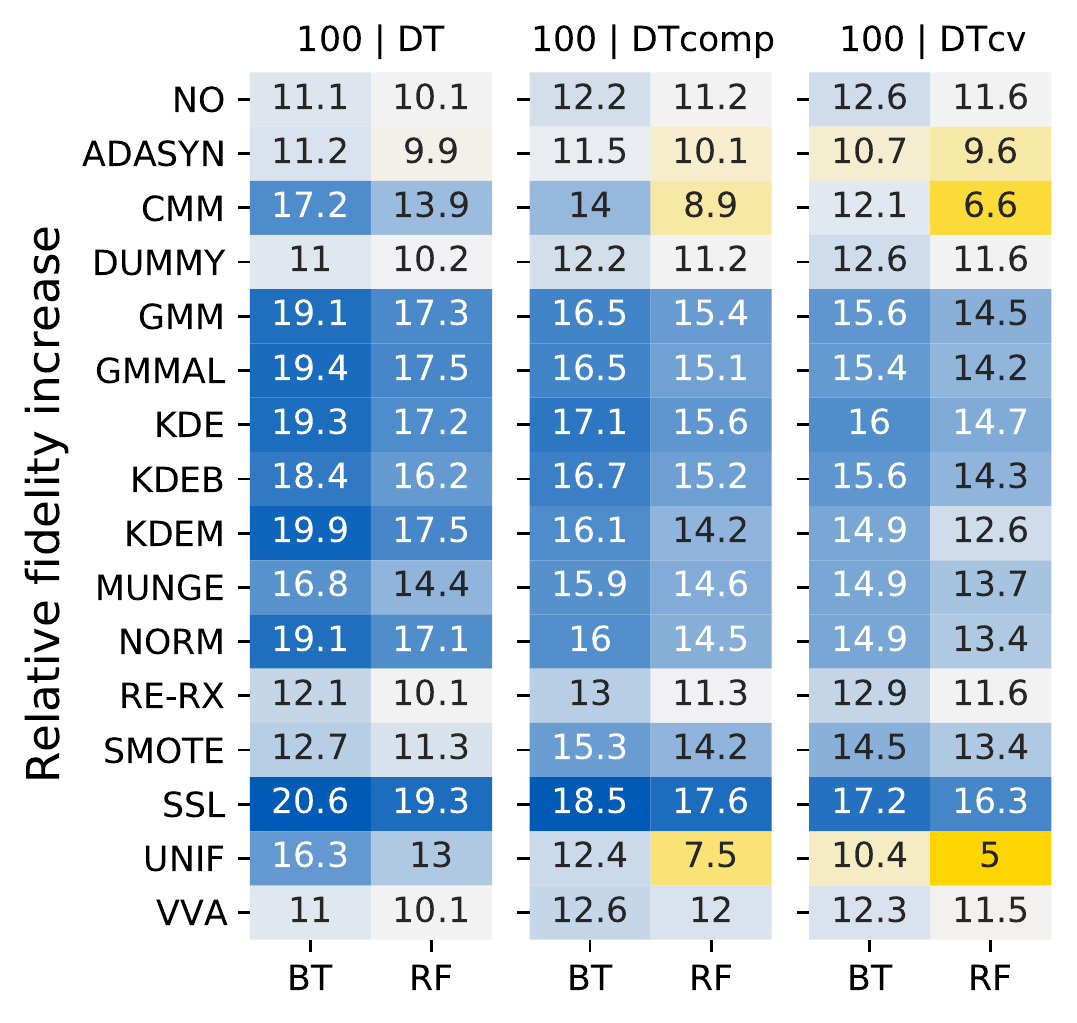}}
	\quad
	{\includegraphics[width=0.40\columnwidth]{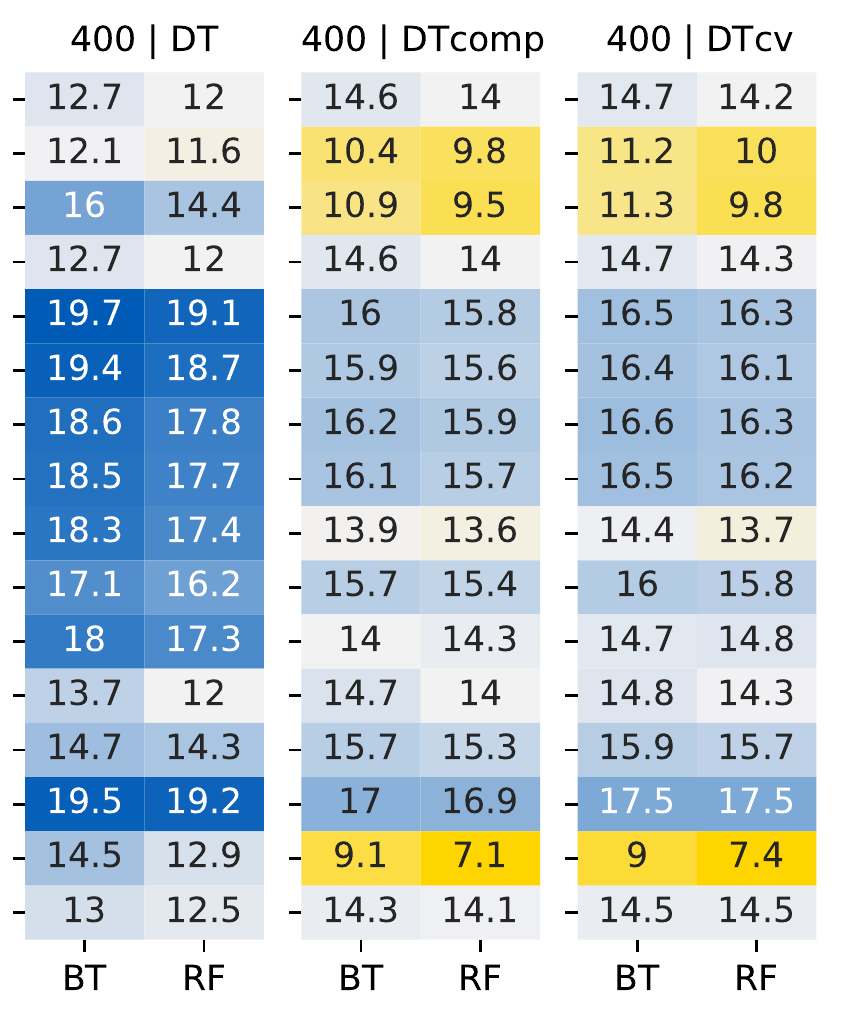}} 
	\caption{Decision tree. Average quality increase for $N=100$ (left) and $N=400$ (right)}
	\label{fig:dt_qual}
\end{figure}

Figures~\ref{fig:dt_nle}--\ref{fig:dt_qual} present the results for decision trees. 
Observe the importance of Requirements \textit{R5} and \textit{R6}: 
While almost all generators improve \texttt{DT} accuracy, for $N=400$, \DTcomp, and \DTcv with \CMM, \KDEM, \NORM, \ADASYN, \UNIF, and \VVA tend to perform worse than the baseline. 
The reason is two-fold.
First, limiting the number of leaves or optimizing it results in a stronger baseline: The accuracy with generator \NO for \DTcomp and \DTcv is higher than respective values for \DT.
Second, some generators only improve the model \wb if it consists of many rules. 

Table~\ref{tab:dt_wdl} reports the average relative accuracy increase of black box models and wins/\allowbreak{draws}/\allowbreak{losses} of \PRELIM with \KDE and decision tree over the baseline for different $N$ and black box models. %
One can compare the values from the column ``bb'' to the relative accuracy increase plots on Figure~\ref{fig:dt_qual}. Accuracy of \wb model obtained from \PRELIM with \KDE is 20--60\% closer to that of \bb model than accuracy of \wb model trained solely on $D^\textit{tr}$. %

So we found that \PRELIM with \KDE, \KDEB, \MUNGE, \SMOTE improves accuracy and fidelity of \DTcomp, \DTcv and yields models with almost the same number of leaves. %
From these generators, only \KDEB has been proposed for pedagogical rule extraction; but its ability to increase the accuracy of decision trees has not been studied before. %

\begin{table}[t]
	\centering
	\caption{Accuracy of decision trees and classification rules, WRAcc of subgroups. Wins/draws/losses of \PRELIM with \KDE versus \NO. The column ``\bb'' reports relative accuracy increase on $D^\textit{test}$ of \bb model learned with respective \BB.}
	\label{tab:dt_wdl}
	\begin{tabular}{c|c|c|c|c|c|c|c|c}\toprule
		\BB &   $N$  & \bb & \DTcomp & \DTcv & {\IREP} & {\RIPPER} & {\BI} & {\PRIM}\\ \hline
		\BT & 100 & 11  & 570/17/161 & 464/104/180 & 594/0/154 & 550/4/194 & 482/34/232 & 562/2/184 \\
		\BT & 400 & 14.4  & 527/4/219 & 460/90/200 & 554/2/194 & 513/5/232 & 504/34/212 & 511/4/235 \\
		\RF & 100 & 11.7  & 578/6/164 & 483/78/187 & 597/2/149 & 542/3/203 & 469/8/271 &	522/0/226 \\
		\RF & 400 & 14.5  & 540/1/209 & 479/64/207 & 535/0/215 & 490/2/258 & 468/7/275 &	452/4/294 \\ \hline
	\end{tabular}
\end{table}

\paragraph{Classification Rules.}

\begin{figure}[t]
	\centering
	\subfloat[Relative accuracy increase]
	{\includegraphics[width=0.36\columnwidth]{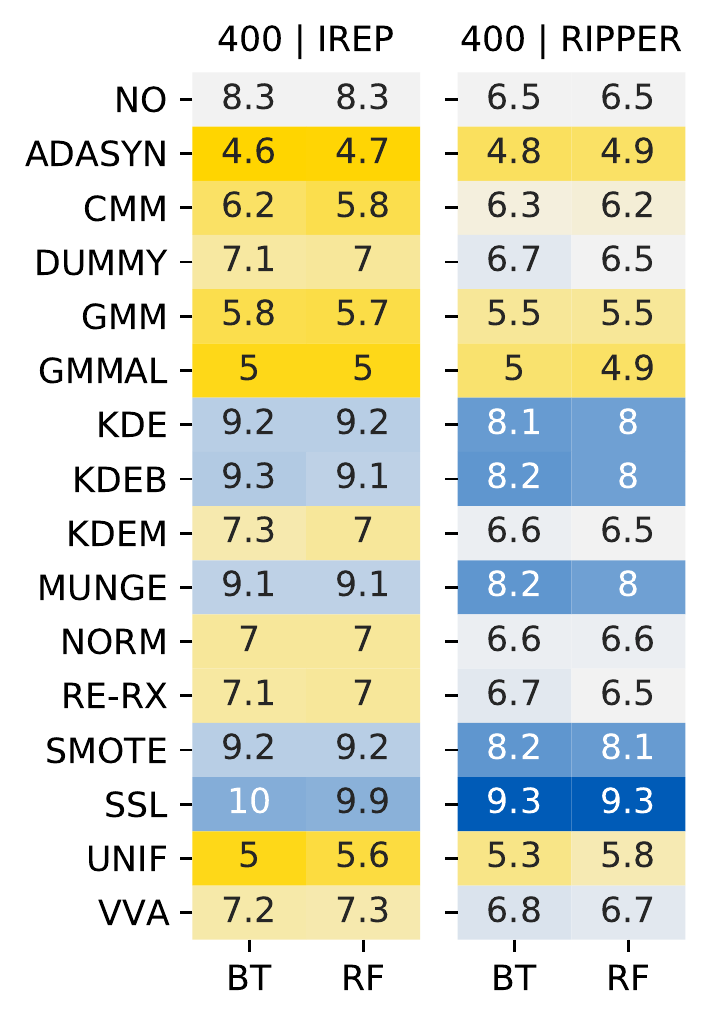}}
	\quad
	\subfloat[Rel. fidelity increase]
	{\includegraphics[width=0.28\columnwidth]{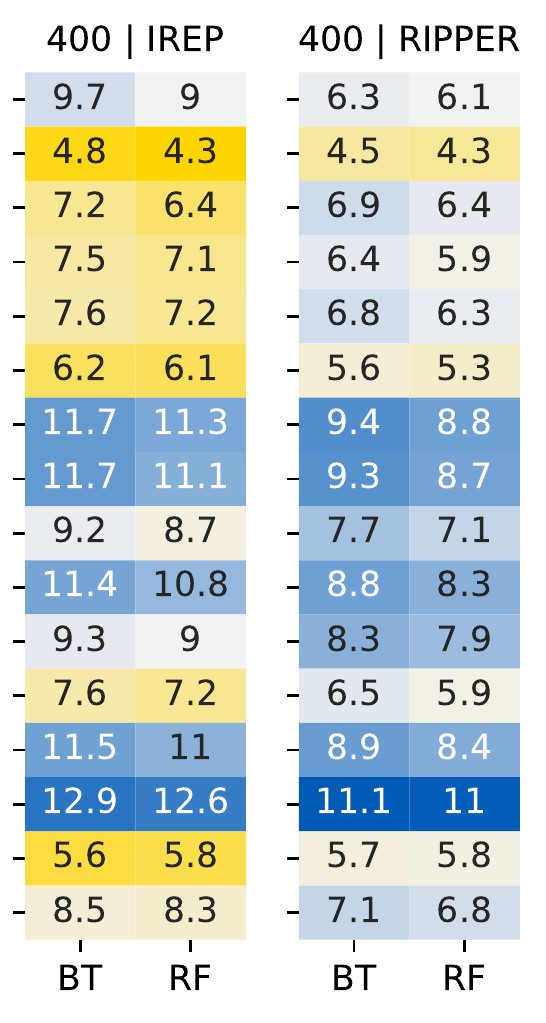}}
	\quad
	\subfloat[Number of rules]
	{\includegraphics[width=0.28\columnwidth]{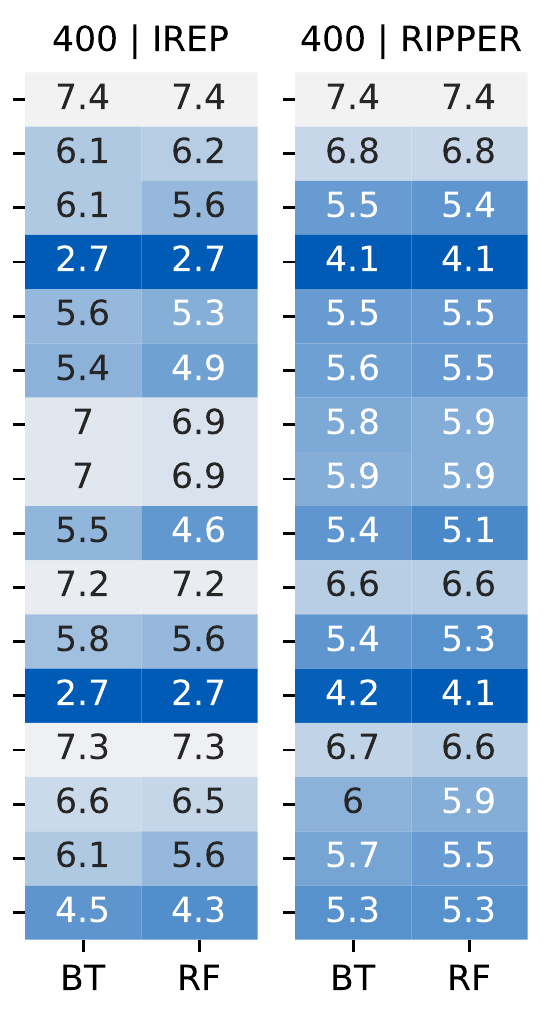}}
	\caption{Classification rules. $N=400$.}
	\label{fig:rules_qual}
\end{figure}

Figure~\ref{fig:rules_qual} presents the results for classification rules, and %
Table~\ref{tab:dt_wdl} lists \IREP and \RIPPER wins/\allowbreak{draws}/\allowbreak{losses} of \PRELIM with \KDE over the baseline %
regarding the relative increase in accuracy. As with decision trees, \KDE, \KDEB, \MUNGE, and \SMOTE generators in \PRELIM increase the accuracy of the rules from \IREP and \RIPPER.

\paragraph{Subgroup Discovery.}

\begin{figure}[t]
	\centering
	\subfloat[WRAcc, $N=100$]
	{\includegraphics[width=0.355\columnwidth]{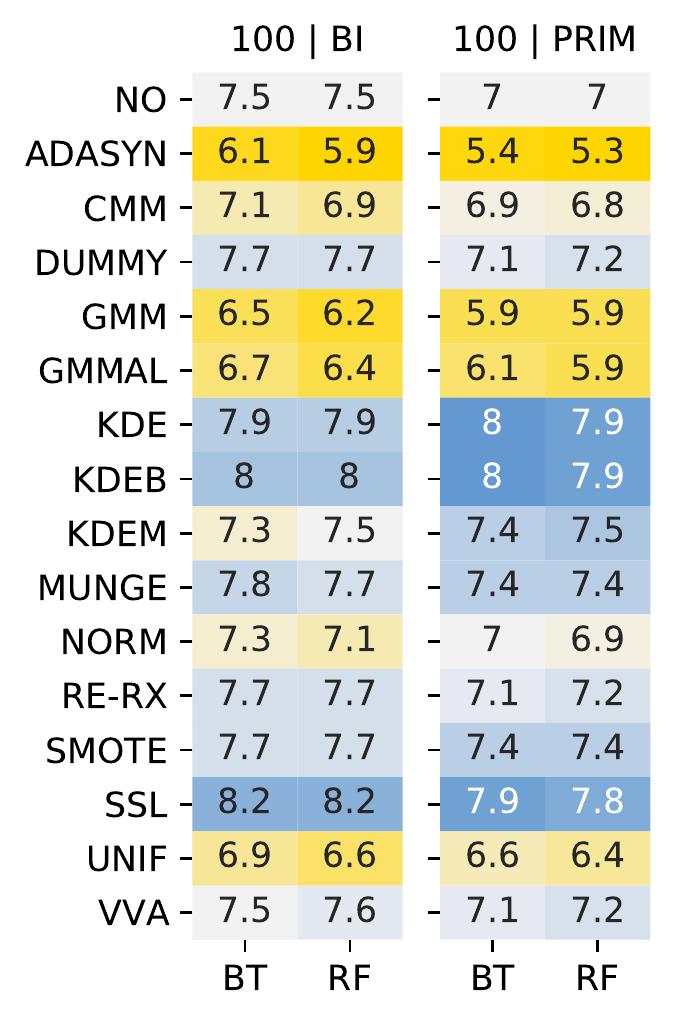}}
	\quad
	\subfloat[WRAcc, $N=100$]
	{\includegraphics[width=0.29\columnwidth]{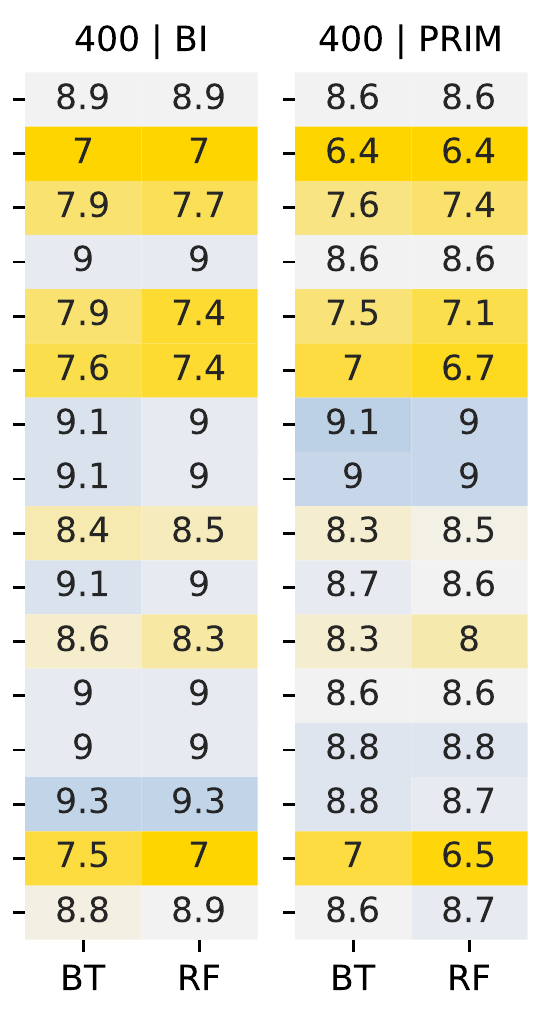}}
	\quad
	\subfloat[``Interpretability'']
	{\includegraphics[width=0.29\columnwidth]{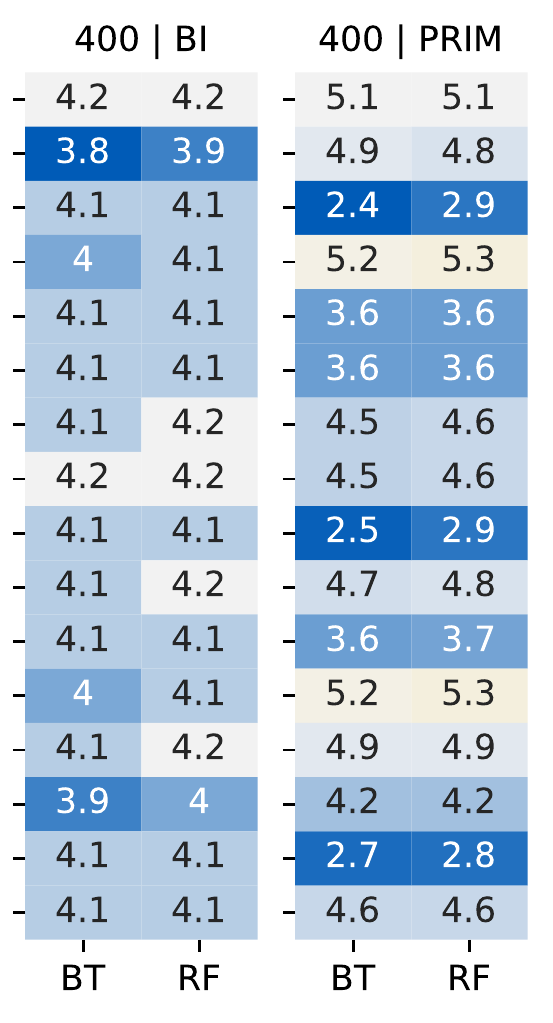}}
	\caption{Subgroup discovery methods.}
	\label{fig:sd_qual}
\end{figure}

Figure~\ref{fig:sd_qual} lists the results for subgroup discovery~--- WRAcc for $N=\{100,400\}$ and the number of features in the rule antecedent (``interpretability'') for $N=400$.
As before, the quality increase is more prominent for smaller datasets~$D^\textit{tr}$. 
Although the WRAcc improvement from 8.9\% to 9.1\% for \KDE and $N=400$ may seem moderate, it is even more significant than that for $N=100$. 
This is visible from Table~\ref{tab:dt_wdl} that reports wins/\allowbreak{draws}/\allowbreak{losses} of \PRELIM with the \KDE generator over the baseline for different $N$ and black-box models concerning WRAcc. One sees that with \SD algorithms as well, \KDE, \KDEB, \MUNGE, and \SMOTE generators in \PRELIM on average demonstrate superior behavior. 

\paragraph{Semi-Supervised Learning.}
The \SSL generator emulates a semi-supervised setting where unlabeled data is available. 
By design, the number of points \SSL outputs in our experiments never exceeds $L$ in other generators except for \DUMMY, \RERX, and \VVA. 
However, white-box models obtained with \SSL tend to be more accurate than those learned by \PRELIM with \KDE. 
This shows that one can use \PRELIM in a semi-supervised setting %
for learning interpretable ML models %
and illustrates the importance of a \PRELIM generator to model joint feature distributions accurately.

\paragraph{Private Data Sharing with \PRELIM.} %
Sometimes one cannot share actual data due to privacy restrictions. One possibility is to share an artificial dataset with similar properties. 
One measure of the quality of the dataset shared is the accuracy of a decision tree learned from this dataset~\cite{DBLP:journals/corr/abs-1811-11264}. 
We have run experiments using only $D^\textit{new}$ (cf.\ Algorithm~\ref{alg:PRELIM}) instead of $D^\textit{tr}\cup D^\textit{new}$ to learn decision trees, with almost the same results.
Thus, one can use \PRELIM %
\KDE + \BT configuration in particular, to preserve privacy by creating artificial data $D^\textit{new}$ of high quality.

%% file: sections/7_Discussion_and_FW.tex
\section{Future Research}
\label{section:future}

While \PRELIM already is quite general, %
we intend to broaden its scope further, as follows.
\PRELIM with subgroup discovery has benefited from using class probability scores instead of class predictions; hence we plan to extend the \PRELIM methodology to use class probabilities with decision trees and classification rules. 
We also plan to test \PRELIM with regression and multiclass classification tasks. 
In the future, we want to extend our framework so that it can handle categorical data and data with mixed features.
We also plan to enrich \PRELIM with other ways to generate new data (e.g., \cite{JSSv074i11,DBLP:journals/jmlr/LiuXGGLW11}) 
and experiment with different approaches to set the hyperparameter values of generators~\cite{DBLP:journals/csda/ZhangKH06}. 
Finally, we want to compare \PRELIM with decompositional rule extraction techniques, cf.~\cite{DBLP:conf/ecml/AsscheB07,DBLP:journals/isci/SagiR21}.

%% file: sections/8_Conclusion.tex
\section{Conclusions}
\label{section:conclusion}

Interpretable white-box models assist many high-stakes decisions. %
In this paper, we deal with the problem of learning accurate white-box models from small datasets. 
Pedagogical rule extraction methods are useful to this end.
They use a black-box model %
to create an augmented dataset, used to learn interpretable models.  
We formulate requirements on these methods and their evaluation and show that existing literature does not sufficiently address them.
Our framework, \PRELIM, allows plugging in various pedagogical rule extraction methods %
to improve learning white-box models from small datasets. 
We have conducted exhaustive experiments that take the requirements into account. %
The experiments show that most existing pedagogical rule extraction methods do not consistently improve the quality of white-box models.
We propose several configurations of \PRELIM,~--- those using \KDE, \KDEB, \MUNGE, and \SMOTE generators,~--- that in turn achieve a consistently higher accuracy of white-box models. %
These configurations also explain complex models better than the competitors and can be used for private data sharing.

%% file: sections/Appendix.tex
\section{White-Box Models}
\label{app:wbmodels}

\paragraph{Decision Trees.}
A decision tree is a classification model in the form of a directed graph consisting of nodes and arcs. Nodes with no direct successors are leaves, the node with no direct predecessor is the root, all others are internal nodes. In the usual case of a binary tree, each node except root has exactly one direct predecessor, and each non-leaf node has two direct successors. Non-leaf nodes contain a test on a feature value, and arcs show the next node to consider depending on if the test is passed or not. Leaf nodes contain the class predicted for the examples reaching it.
Decision tree learning algorithms differ in heuristics used to learn tests for non-leaf nodes and to prune a tree. We use a version of a popular decision tree learning algorithm, \CART~\cite{DBLP:books/wa/BreimanFOS84}\footnote{\url{https://scikit-learn.org/stable/modules/tree.html}}.

\paragraph{Classification Rules.}
A classification rule model consists of \texttt{if-then} rules. The part between \texttt{if} and \texttt{then} is the antecedent of a rule and contains conjunction of tests on different features. The part after \texttt{then} is the rule consequent; it contains a predicted class label. In contrast to decision trees that produce mutually exclusive rules, classification rules can overlap. Usually, rules are ordered to form a so-called decision list~\cite{DBLP:series/cogtech/FurnkranzGL12}, i.e., \texttt{if-then-else} rules. To predict a class label with a decision list, one uses the first rule that fires. If no rule fires, the model assigns a ``default'' class.
With \PRELIM, we use a popular rule learning algorithm, {\IREP}~\cite{DBLP:conf/icml/FurnkranzW94}, and its successor {\RIPPER}~\cite{DBLP:conf/icml/Cohen95}. The main differences between \RIPPER and \IREP are in stopping criterion, pruning heuristic, and optimization technique of the latter; see~\cite{DBLP:series/cogtech/FurnkranzGL12} for details.

\paragraph{Subgroup Discovery.}
In contrast to classification rules, subgroup discovery methods focus on the properties of individual rules~\cite{DBLP:series/cogtech/FurnkranzGL12}. Weighted relative accuracy (WRAcc) commonly measures individual rule quality.
In \PRELIM, we use \BI~\cite{DBLP:conf/icdm/MampaeyNFK12} and \PRIM~\cite{DBLP:journals/sac/FriedmanF99} subgroup discovery algorithms that can work with continuous features. \PRIM starts with defining a target function (WRAcc in this paper). It then finds a rule by cutting off a small share $\alpha$ of examples to maximize WRAcc. \BI greedily maximizes WRAcc considering one feature at a time~\cite{DBLP:conf/sigmod/ArzamasovB21}.

\section{Valley-Valley Approximation Generator}
\label{app:vva}

The \texttt{vva} generator creates new points as follows.
\begin{enumerate}[noitemsep]
	\item Take $N_v$ train data points with the most uncertain predictions of $\bb$;
	\item for each point, find its nearest neighbor with a different predicted class;
	\item sort the resulting pairs of nearest neighbors according to distances between them, starting from smaller distances;
	\item for each pair of points, generate a new point on the line connecting them;
	\item repeat the previous step several times to obtain $L$ new points.
\end{enumerate}
The ratios $N_v/N$ and $L/N$ are hyperparameters of \texttt{vva}.

\section{Experimental Setup. Additional Details}
\label{app:experimental_setup}

\subsection{Datasets}
\label{app:datasets}

Table~\ref{tab:datasets} lists the datasets together with their characteristics after the pre-processing and corresponding references. Here $\DSIZE$ is the number of rows, $M$ is the number of features, ``Pos\_class''~--- our criterion to assign the class label ``1'' in datasets for multi-class classification or regression, $\lvert D^+\rvert$~--- the number of rows belonging to class ``1''.

\subsection{Implementation Details}

We implement the experiments in Python~\cite{10.5555/1593511}.
For decision trees and random forests, we rely on scikit-learn~\cite{pedregosa2011scikit}. Boosted trees are from XGBoost~\cite{DBLP:conf/kdd/ChenG16}. Classification rules are from the ``wittgenstein'' repository\footnote{\url{https://github.com/imoscovitz/wittgenstein}}; we fixed several bugs in it. We implemented subgroup discovery methods according to~\cite{DBLP:conf/sigmod/ArzamasovB21}, and generators according to the descriptions in respective papers.

\subsection{Hyperparameters Used in the Experiments}
\label{app:hpo}

\paragraph{\PRIM.}
For \PRIM, we optimize $\alpha$ (cf. Section~\ref{app:wbmodels}) by choosing its value from the set \{0.03, 0.05, 0.07, 0.1, 0.13, 0.16, 0.2\}. 

\paragraph{Black-Box Models.}
For random forest, we select max\_features from $[2, \sqrt{M}, M]$.
For boosted trees, we do random search in the following hyperparameter space: $\textrm{n\_estimators}$ is integer from the range $[10,990]$, 
$\textrm{learning\_rate}\in[0.0001,0.2]$, $\textrm{gamma}\in[0,0.4]$
$\textrm{max\_depth} = 6$, $\textrm{subsample}\in[0.5,1]$, the other hyperparameters are equal to their default values as provided in implementation. This is similar to what~\cite{DBLP:conf/gecco/OrzechowskiCM18} does.

\paragraph{Generators.}

For the \GMM generator, we adjust hyperparameters using the Bayesian information criterion~\cite{schwarz1978estimating}. We optimize the covariance matrix structure by choosing it from all possible structures offered by the implementation and the number of components $k$~--- from $\{1,\dots,29\}$. With \GMMAL, we proceed similarly, except the covariance structure is restricted to diagonal. To set the bandwidth of \KDEM, we use Silverman's rule of thumb~\cite{DBLP:books/sp/Silverman86}:
$$
	h_X=\frac{0.9A}{N^{1/5}},\quad A=\min\left(\sqrt{\textrm{Var}(X)},\frac{\textrm{IQR}(X)}{1.349}\right).
	\label{eq:bandwidth1D}
$$
Here $N$ is the number of observations, as before; $\textrm{Var}(X)$ is the sample variance estimate of the feature $X$, $\textrm{IQR}(X)$ is the feature's interquartile range.
For \KDE (multidimensional version density estimate), we set the bandwidth matrix to
$$
	H=\frac{I_M}{M}\sum_{i=1}^{M}h_{X_i}.
	\label{eq:bandwidthND}
$$
Here $M$ is the number of features; $I_M$~--- the $M\times M$ identity matrix; $h_{X_i}$~--- the bandwidth calculated for the feature $X_i$ with~(\ref{eq:bandwidth1D}).
In line with~\cite{DBLP:journals/tnn/SchmitzAG99}, we set $r$ in \KDEB equal to the average distance of points in $D^\textit{tr}$ to their 10-th nearest neighbors. For the \CMM generator, we use the same random forest model as the one used as $\bb$ in the respective \PRELIM instantiation. Following~\cite{DBLP:journals/tnn/FortunyM15}, for \VVA, we set $N_v/N=0.2$ and select $L/N$ from $[0,2.5]$ via cross-validation.

For \SMOTE and \ADASYN generators, we use the default hyperparameters from their implementations. Specifically, we set the number of nearest neighbors $k=5$. \ADASYN fails in case of the absence of points from $P^\textit{new}$ in the $k$-neighborhood of a point from $D^\textit{tr}$. In this case, we try to increase $k$ for \ADASYN gradually; if it does not help, we use \SMOTE instead of \ADASYN in the respective experiment.

The work~\cite{DBLP:conf/kdd/BucilaCN06} does not recommend particular values for hyperparameters $P$ and $s$ for \MUNGE. After preliminary experiments, we set their values to $P=0.5$ and $s=5$.
\SSL returns $L$ points from $D^\textit{test}$ and leaves only the remaining examples for testing.

\begin{table}
	\centering
	\caption{Datasets.}
	\label{tab:datasets}
	\begin{tabular}{lccccc}
		\toprule
		Name        & $\DSIZE$     & $M$   & $\lvert D^+\rvert/\DSIZE$ & Pos\_class & Ref.                               \\ \hline
		anuran      & 7195  & 21  & 0.3  & Hylidae                 & \cite{Dua:2019}                             \\
		avila       & 20867 & 10  & 0.41 & A                       & \cite{DBLP:journals/eaai/StefanoMFF18}      \\
		bankruptcy  & 4885  & 64  & 0.02 & 1                       & \cite{zikeba2016ensemble}                   \\
		ccpp        & 9568  & 4   & 0.45 & PE $> 455$                & \cite{kaya2012local}                        \\
		cc          & 30000 & 14  & 0.22 & 1                       & \cite{DBLP:journals/eswa/YehL09a}           \\
		clean2      & 6598  & 166 & 0.15 & 1                       & \cite{romano2021pmlb}                       \\
		dry         & 13611 & 16  & 0.26 & DERMASON                & \cite{DBLP:journals/cea/KokluO20}           \\
		ees         & 14980 & 14  & 0.55 & 1                       & \cite{Dua:2019}                             \\
		electricity & 45312 & 7   & 0.42 & UP                      & \cite{DBLP:journals/sigkdd/VanschorenRBT13} \\
		gas         & 13910 & 128 & 0.18 & 1                       & \cite{vergara2012chemical}                  \\
		gt          & 19020 & 10  & 0.65 & g                       & \cite{Dua:2019}                             \\
		higgs21     & 98049 & 17  & 0.53 & 1                       & \cite{baldi2014searching}                   \\
		higgs7      & 98049 & 7   & 0.53 & 1                       & \cite{baldi2014searching}                   \\
		htru        & 17898 & 8   & 0.09 & 1                       & \cite{lyon2016fifty}                        \\
		jm1         & 10880 & 21  & 0.19 & True                    & \cite{Sayyad-Shirabad+Menzies:2005}         \\
		ml          & 6118  & 51  & 0.42 & 1                       & \cite{DBLP:journals/jar/BridgeHP14}         \\
		nomao       & 34465 & 69  & 0.29 & 1                       & \cite{candillier2012design}                 \\
		occupancy   & 20560 & 6   & 0.23 & 1                       & \cite{candanedo2016accurate}                \\
		parkinson   & 5875  & 16  & 0.44 & motor\_UPDRS $> 23$        & \cite{DBLP:journals/tbe/TsanasLMR10}        \\
		pendata     & 10992 & 16  & 0.1  & 1                       & \cite{Dua:2019}                             \\
		ring        & 7400  & 20  & 0.5  & 1                       & \cite{romano2021pmlb}                       \\
		saac2       & 15533 & 21  & 0.49 & 1                       & \cite{DBLP:journals/sigkdd/VanschorenRBT13} \\
		seizure     & 11500 & 178 & 0.2  & 1                       & \cite{andrzejak2001indications}             \\
		sensorless  & 58509 & 48  & 0.09 & 1                       & \cite{Dua:2019}                             \\
		seoul       & 8760  & 7   & 0.36 & Rented Bike Count $> 800$ & \cite{DBLP:journals/comcom/EPC20}           \\
		shuttle     & 58000 & 9   & 0.79 & 1                       & \cite{romano2021pmlb}                       \\
		stocks      & 96320 & 21  & 0.51 & 1                       & \cite{DBLP:journals/sigkdd/VanschorenRBT13} \\
		sylva       & 14395 & 20  & 0.06 & 1                       & \cite{DBLP:journals/sigkdd/VanschorenRBT13} \\
		turbine     & 36733 & 9   & 0.29 & NOX $> 70$               & \cite{kaya2019predicting}                   \\
		wine        & 4898  & 11  & 0.45 & quality $= 6$            & \cite{DBLP:journals/dss/CortezCAMR09}  \\	\hline  
	\end{tabular}
\end{table}